\documentclass[final]{cvpr}

\usepackage{times}
\usepackage{epsfig}
\usepackage{graphicx}
\usepackage{amsmath}
\usepackage{amssymb}

\usepackage{amsmath}
\usepackage{subcaption}
\usepackage{algorithm}
\usepackage{algorithmic}

\usepackage[pagebackref=true,breaklinks=true,colorlinks,bookmarks=false]{hyperref}

\def\cvprPaperID{7627} 
\def\confYear{CVPR 2021}

\begin{document}

\title{Evolving Neural Architecture Using One Shot Model}

\author{Nilotpal Sinha\\
National Chiao Tung University\\
Hsinchu City, Taiwan\\
{\tt\small nilotpalsinha.cs06g@nctu.edu.tw}
\and
Kuan-Wen Chen \\
National Chiao Tung University\\
Hsinchu City, Taiwan\\
{\tt\small kuanwen@cs.nctu.edu.tw}
}

\maketitle

\begin{abstract}
Neural Architecture Search (NAS) is emerging as a new research direction which 
has the potential to replace the hand-crafted neural architectures designed for 
specific tasks. Previous evolution based architecture search requires high computational
resources resulting in high search time. In this work, we propose a novel way of applying a
simple genetic algorithm to the NAS problem called EvNAS \emph{(Evolving Neural Architecture
using One Shot Model)} which reduces the search time significantly while still achieving
better result than previous evolution based methods. The architectures are represented by
using the architecture parameter
of the one shot model which results in the weight sharing among the architectures
for a given population of architectures and also weight inheritance from one
generation to the next generation of architectures. We propose a decoding technique for
the architecture parameter which is used to divert majority of the gradient
information towards the given architecture and is also used for improving the
performance prediction of the given architecture from the one shot model
during the search process. Furthermore, we use the accuracy of the partially trained
architecture on the validation data as a prediction of its fitness in order to reduce the
search time. EvNAS searches for the architecture on the \textit{proxy} dataset i.e. CIFAR-10
for 4.4 GPU day on a single GPU and achieves top-1 test error of 2.47\%
with 3.63M parameters which is then transferred to CIFAR-100 and ImageNet
achieving top-1 error of 16.37\% and top-5 error of 7.4\% respectively. All of
these results show the potential of evolutionary methods in solving the architecture search
problem.
\end{abstract}

\section{Introduction}
Convolutional neural networks have been instrumental in solving various 
problems in the field of computer vision. However, the network designs were mainly done by humans
(like AlexNet \cite{krizhevsky2012imagenet}, ResNet \cite{he2016deep}, DenseNet
\cite{huang2017densely}, VGGNet \cite{simonyan2014very}) on the basis of their intuition and
understanding of the specific problem. This has led to the growing interest in the automated
search of neural architecture called \textit{Neural Architecture Search} (NAS)
\cite{elsken2018neural}\cite{zoph2016neural}\cite{pmlr-v80-pham18a}. NAS has shown some promising
results in the field of computer vision but most of these methods demand a considerable amount of
computational power. For example, obtaining the  state-of-the-art architecture for CIFAR-10
required 3150 GPU days  of evolution \cite{real2019regularized} and 1800 GPU days of reinforcement
learning (RL)  \cite{zoph2018learning}. This can be mainly attributed to the evaluation of the
architectures during the search process of the different NAS methods because most NAS methods
train each architecture individually for certain number of epochs in order to evaluate its
performance on the validation data. Recent works
\cite{pmlr-v80-pham18a}\cite{bender2019understanding} have reduced the search time by weight
sharing among the architectures. DARTS \cite{liu2018darts2} further improves upon the scalability
by relaxing the search space to a continuous space in order to use gradient descent for optimizing
the architecture. But these gradient based methods are highly dependent on the search space and
they tend to overfit to operations in the search space that lead to faster gradient
descent \cite{Zela2020Understanding}.

In this work, we propose a method called EvNAS (\textit{Evolving Neural 
Architecture using One Shot Model}) which involves evolving a convolutional neural
network architecture with weight sharing among the architectures in the population for the
image classification task. The work is inspired in part by the representation used for
the architectures of the network in DARTS \cite{liu2018darts2} and a random
search \cite{li2019random} using the same representation as DARTS which
achieved a competitive result on the CIFAR-10 dataset. By replacing the idea of using
a random search with a genetic algorithm on the representation used in DARTS, we
introduce a directional component to the otherwise directionless random search
\cite{li2019random} through the use of crossover with tournament selection and elitism. The
stochastic nature of the algorithm also ensures that the algorithm does not get stuck in the
local minima.

The objective of the paper is to show how to apply a simple genetic algorithm to the
neural architecture search problem while reducing the high search time associated with 
the evolution based search algorithm. Our experiments (Section ~\ref{experiments}) involves
neural architecture search on a proxy dataset i.e. CIFAR-10 which achieves the
test error of 2.47\% with 3.63M parameters on this dataset while using minimal computational
resources (4.4 GPUs day on a single GPU). The discovered architecture is then transferred to
CIFAR-100 and ImageNet achieving top-1 error of 16.37\% and top-5 error of 7.4\%
respectively.

Our contributions can be summarized as follows:
\begin{itemize}
    \item We introduce a novel method of applying a simple genetic algorithm to the NAS problem with
    reduced computational requirements.
    \item We propose a decoding technique for each architecture in the population which diverts a
    majority of the gradient information to the current architecture during the training phase and is
    used to calculate the fitness of the architecture from the one shot model during the fitness
    evaluation phase.
    \item We propose a crossover operation that is guided by the predicted fitness of the partially
    trained architectures of the previous generation and does not require keeping track of the
    ancestors of the parent architectures.
    \item We achieved remarkable efficiency in the architecture search achieving
    test error of 2.47\% with 3.63M parameters on CIFAR-10 and showed that the architecture
    learned by EvNAS is transferable to CIFAR-100 and ImageNet.
\end{itemize}

\begin{figure*}
  \centering
  \begin{subfigure}{0.34\linewidth}
    \includegraphics[width=\linewidth,scale=0.5]{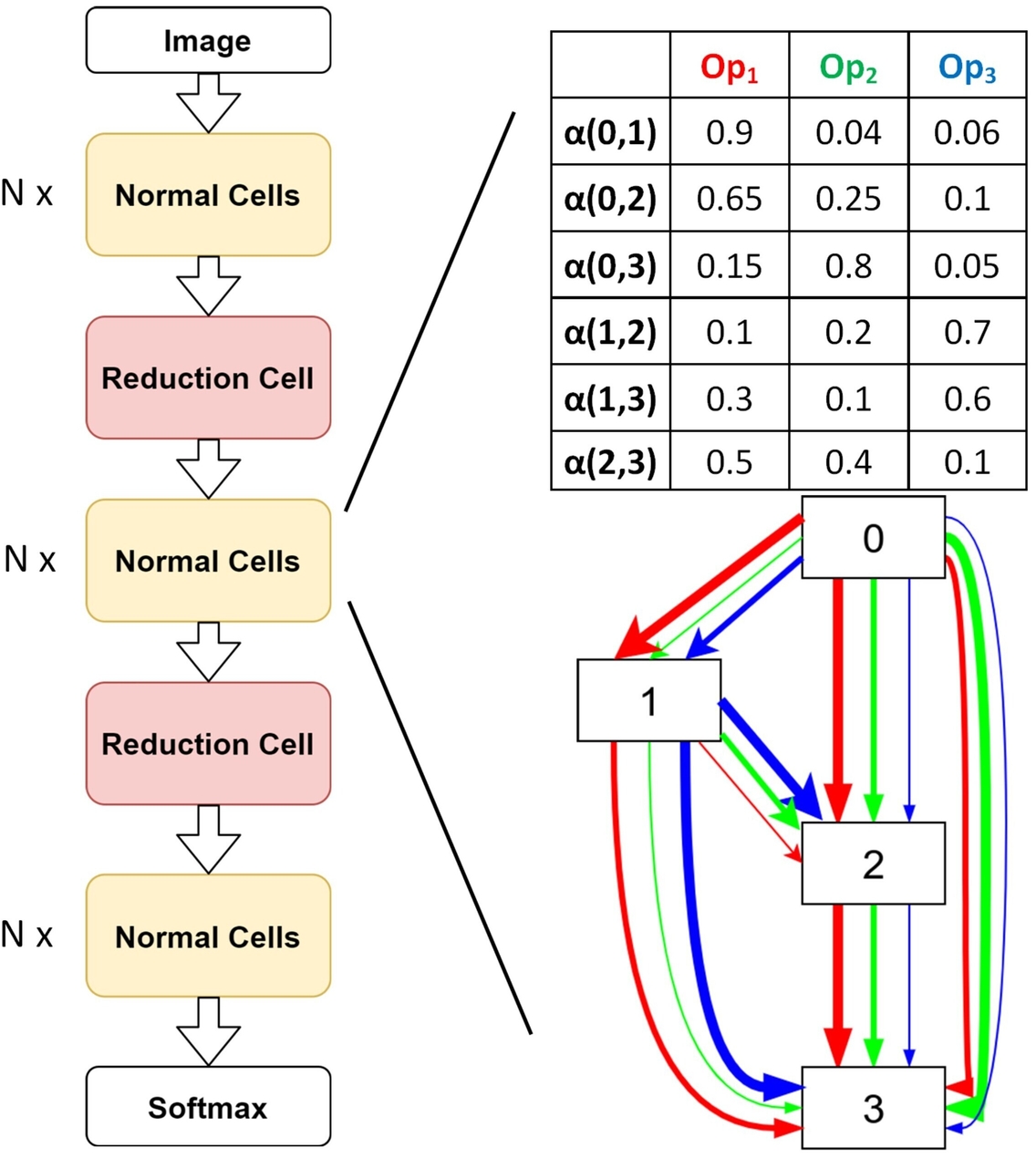}
    \caption{}
    \label{subfig:darts}
  \end{subfigure}
  \quad
  \begin{subfigure}{0.18\linewidth}
    \includegraphics[width=\linewidth]{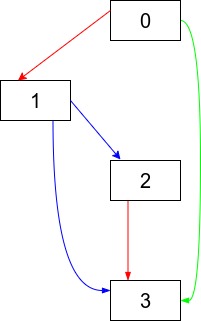}
    \caption{}
    \label{subfig:discrete_arch}
  \end{subfigure}
  \quad
  \begin{subfigure}{0.38\linewidth}
    \includegraphics[width=\linewidth]{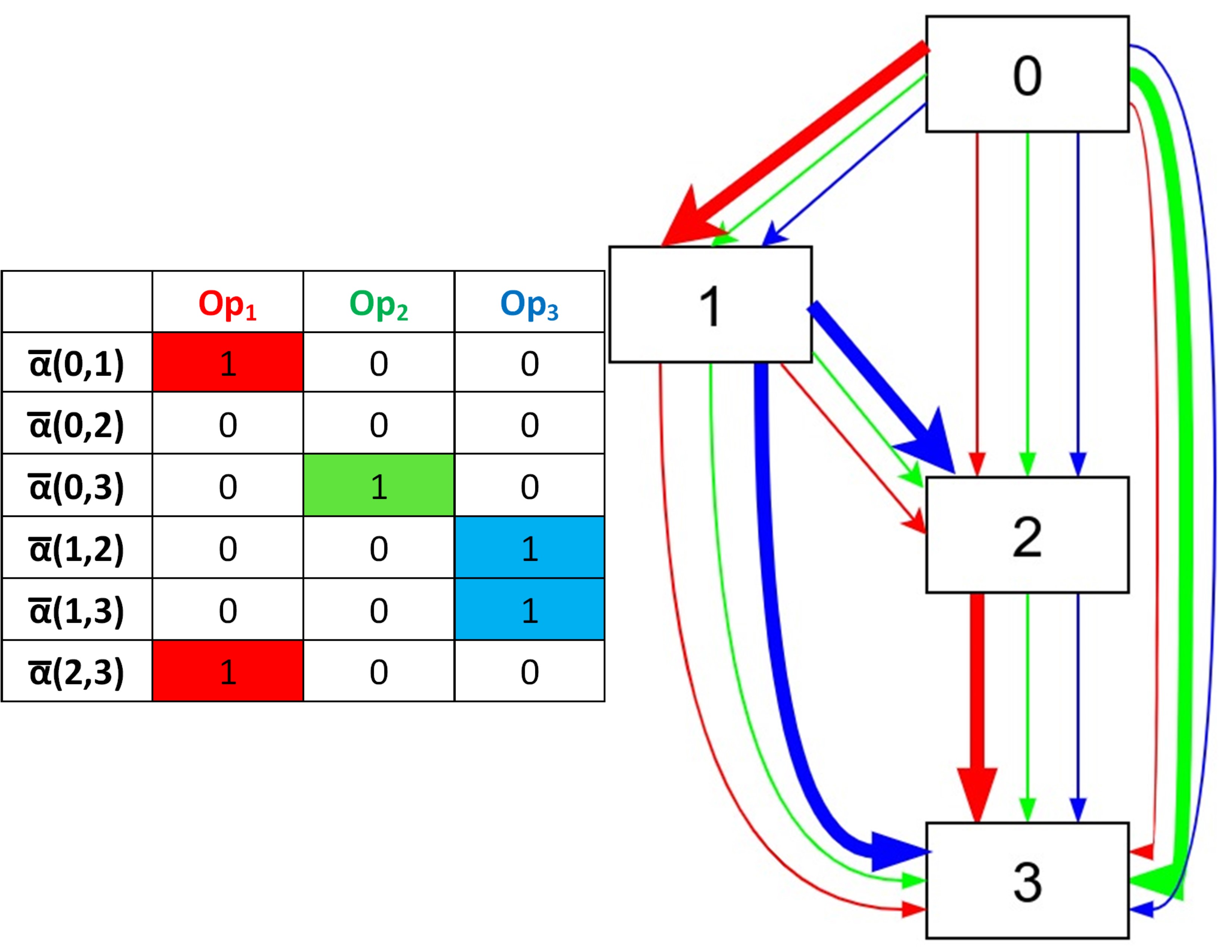}
    \caption{}
    \label{subfig:decoded_arch}
  \end{subfigure}
  \caption{The process of decoding the architecture parameter, $\alpha$. Better viewed in
  color mode. Here, we consider three operations in the operation space.
  (a) One shot model and its representation with arrows between the nodes
  representing all the operations in the search space,
  (b) Discrete architecture, $arch_{dis}$, derived from $\alpha$,
  (c) Decoded architecture, $\bar{\alpha}$, created using $arch_{dis}$.
  The thickness of the arrow is proportional to the weight given to an operation.}
  \label{fig:arch_represent}
\end{figure*}

\section{Related Work}
Automated Neural Architecture Search is an alternative to the  hand-crafted architectures where the
machine designs the best suited architecture for a specific problem. Several search methods have
been proposed to explore the space of neural architectures, such as evolutionary algorithm (EA)
\cite{real2019regularized}\cite{real2017large}\cite{liu2018hierarchical}\cite{xie2017genetic},
reinforcement learning (RL) \cite{zoph2016neural}\cite{zoph2018learning}\cite{pmlr-v80-pham18a},
random search\cite{li2019random} and gradient-based methods
\cite{liu2018darts}\cite{liu2018darts2}\cite{luo2018neural}\cite{chen2019progressive}\cite{Zela2020Understanding}.
These can be grouped into two groups: \textit{gradient-based} methods and \textit{non-gradient based}
methods.

\textbf{Gradient Based Methods:} In these methods, the neural architecture is directly optimized
using the gradient information based on the performance on the validation data. In
\cite{liu2018darts}\cite{liu2018darts2}, the discrete architecture search space is 
relaxed to a continuous search space by using a one shot model and the performance of the
model on the validation data is used for updating the architecture using gradients. This method
reduces the search time significantly but suffers from the overfitting problem wherein the searched
architecture performs very well on the validation data but exhibits poor performance on the test
data. This is mainly attributed to the preference of parameter-less operations during the search
process as it leads to a rapid gradient descent \cite{chen2019progressive}. Many regularizations
have been introduced to tackle the problem such as early stopping \cite{Zela2020Understanding},
search space regularization \cite{chen2019progressive} and architecture refinement
\cite{chen2019progressive}. Contrary to the gradient based methods, the proposed method does not
suffer from the overfitting problem because of the stochastic nature introduced by the mutation
operation.

\textbf{Non-Gradient Based Methods:} These methods include reinforcement learning (RL) and evolutionary algorithm (EA). In RL methods, an agent is trained to generate a neural architecture
through its action in order to maximize the expected accuracy on the validation data. In
\cite{zoph2016neural}\cite{zoph2018learning}, a recurrent neural network (RNN) is used as an agent
which samples neural architectures which are then trained to convergence in order to obtain their
accuracies on the validation data. These accuracies are then used to update the weights of RNN by
using policy gradient methods. Both of these methods suffered from huge computational requirements.
This was improved upon in \cite{pmlr-v80-pham18a}, where all the sampled architectures were forced
to share weights by using a single directed acyclic graph (DAG) resulting in the reduction of
computational resources. Early approaches based on EA such as
\cite{stanley2002evolving}\cite{stanley2009hypercube} optimized both the neural architectures and
the weights of the network which limited their usage to relatively smaller networks. Then, methods
such as \cite{xie2017genetic}\cite{real2019regularized} used evolution to search for the
architecture and gradient descent for optimizing the weights of each architecture which made it
possible to search for relatively large networks. However, this resulted in huge computational
requirements. To speed up the training of each individual architecture, \textit{weight inheritance}
was introduced in \cite{real2017large} wherein a child network inherits the parent networks'
weights. In this work, we used both weight inheritance and weight sharing among the architectures
to speed up the search process.

\section{Methods}
\label{methods}
This section discusses different parts of the proposed algorithm and its relationship to prior works.

\subsection{Representation of Architecture}
The proposed algorithm deals with a population of architectures in each generation during the 
search process. Instead of having a separate model for each architecture in a population
\cite{xie2017genetic}\cite{real2019regularized}, we used a \textit{one shot model} which
treats all the architectures as subgraphs of the supergraph while sharing the weights among
all the architectures. The one shot model is composed of repeatable cells which are stacked
together to form the convolutional network. The one shot model has two types of convolutional
cells: \textit{normal} cell and \textit{reduction} cell. A normal cell uses operations with
stride 1 whereas reduction cell uses operations with stride 2. A cell in the one shot model is
represented by the parameter, $\alpha$ called \textit{architecture parameter}, which
represents the weights of the different operations $op(.)$ in the operation space $O$ (i.e.
search space of NAS) between a pair of nodes. The edge between node $i$ and node
$j$ can be written as:
    \begin{equation}
       f^{(i,j)}(x) = \sum_{op \in O } 
            \frac{exp(\alpha^{i,j}_{op})}
                 {\sum_{op' \in O } exp(\alpha^{i,j}_{op'})} op(x).
    \end{equation}
Where $\alpha^{i,j}_{op}$ refers to the weight of the operation $op$ in the
operation space $O$ between node $i$ and node $j$. The architecture is represented by two 
matrices, one for normal cell and one for reduction cell, where the row represents the edge
between two nodes and the column represents the weights of different operations from the
operation space as shown in Figure~\ref{fig:arch_represent}(a). Please refer to the original
DARTS paper \cite{liu2018darts2} for more technical details.
The design choice results in weight sharing among the architectures in a given population of
architectures. It also results in weight inheritance from one generation of 
architectures to the next generation of architectures i.e. the next generation
architectures are not trained from scratch but inherit the partially trained weights
from the previous generation architectures. All of these ultimately
leads to the reduction of the architecture search time using evolution.

\begin{figure*}[t]
  \centering
  \begin{subfigure}{0.3\linewidth}
    \includegraphics[width=\linewidth]{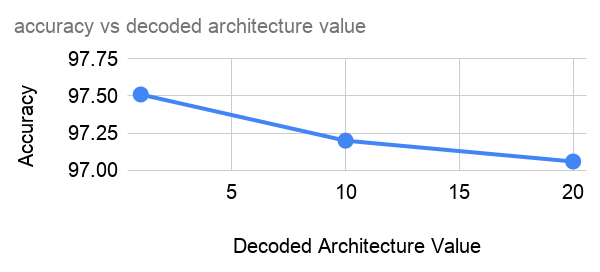}
    \caption{}
    \label{subfig:AccVsArch}
  \end{subfigure}
  \quad
  \begin{subfigure}{0.3\linewidth}
    \includegraphics[width=\linewidth]{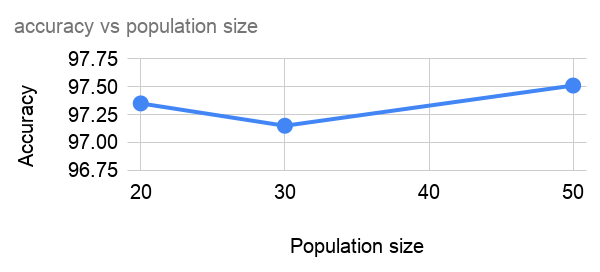}
    \caption{}
    \label{subfig:AccVsPop}
  \end{subfigure}
  \quad
  \begin{subfigure}{0.3\linewidth}
    \includegraphics[width=\linewidth]{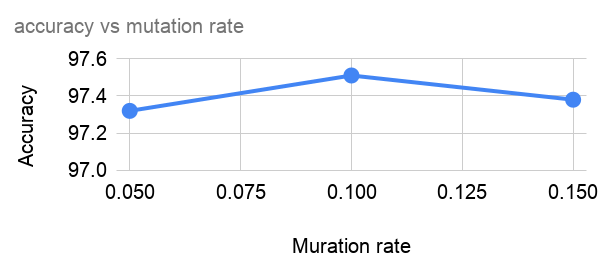}
    \caption{}
    \label{subfig:AccVsMrate}
  \end{subfigure}
  
  \caption{(a) Accuracy vs Decoded architecture value
           (b) Accuracy vs Population size
           (c) Accuracy vs mutation rate}
  \label{fig:exp}
  \end{figure*}

\subsection{Decoding Architecture Parameter}\label{subsec:decode}
\textit{Architecture parameter}, $\alpha$, gives variable weights to the operations in any
particular architecture which results in very noisy estimate of fitness of the architecture. This
results in the algorithm performing marginally better than the random algorithm as discussed in
Section ~\ref{ablation}. We propose a decoding technique, which is a process of giving equal higher
weight to the operations of the actual architecture/subgraph according to the architecture
parameter, $\alpha$ and equal smaller weights to the operations of the other architectures. This
can be thought of as decoding/mapping the genotype, i.e. $\alpha$, to the phenotype, i.e. actual
architecture \cite{eiben2003introduction}. The process has the following two steps:
\begin{itemize}
    \item For any $\alpha$, derive the discrete architecture, $arch_{dis}$, from 
    $\alpha$ as shown in Figure~\ref{fig:arch_represent}(b).
    \item On the basis of the discrete architecture, $arch_{dis}$, create 
    another architecture parameter called \textit{decoded architecture} parameter
    , $\bar{\alpha}$ (as shown in Figure~\ref{fig:arch_represent}(c)), with the following entries:
        \begin{equation}
            \bar{\alpha}^{i,j}_{op} = 
                \begin{cases}
                k, \text{if $op$ between node $i$ and $j$ present in $arch_{dis}$}
                \\
                0,  \text{otherwise}
                \\
                \end{cases}    
        \end{equation}
\end{itemize}
where \textit{k} is an integer. The design ensures that the current architecture according to
$\alpha$ gets a majority of the gradient information to update its parameters while the rest of
the gradient information is distributed equally among all the other architectures to update their
parameters. This results in making sure that the weights of an architecture does not get
co-dependent with the weights of the other architecture due to the weight sharing nature of the
one shot model. It also helps in improving the estimation of the \textit{fitness} of each
architecture in the population, as it gives higher equal weight to that particular architecture
operations while giving lower equal weights to the other architecture operations. This results in
the higher contribution from a particular architecture while very low contribution by other
architectures during the fitness evaluation step of that particular architecture from the one shot
model. This is in contrast to the variable architecture contribution, used in the original DARTS
paper, wherein an architecture is evaluated using $\alpha$ which results in very noisy estimate of
its performance. We empirically find that k = 1 gives a good result and increasing the value of k
from 1 tends to deteriorate the accuracy as shown in Figure \ref{fig:exp}(a).

\subsection{Training and Performance Estimation}
The sharing of the network weights among the architectures in the population, due to the one shot
model representation \cite{liu2018darts2}, helps in exchanging information to the next generation
population, wherein the architectures of the next generation do not start training from scratch.
This can be thought of as child architecture model inheriting the weights of the parent
architecture model, also known as \textit{weight inheritance}. Therefore, instead of the full
training of each architecture in the population from scratch, EvNAS partially trains the inherited
architecture model weights by using the training data. This is done by first copying the
\textit{decoded architecture} parameter, $\bar{\alpha}$, in Section~\ref{subsec:decode}, for the
individual architecture in the population to the one shot model and then training the network for
a certain number of batches of training examples. 

To evaluate the performance of each individual architecture, its \textit{decoded architecture}
parameter, $\bar{\alpha}$, from Section~\ref{subsec:decode}, is first copied to the one shot 
model. The model is then evaluated on the basis of its accuracy on the validation data, 
which becomes the \textit{fitness} of the architecture. Note that the fitness value of each
architecture is a noisy estimate of its true accuracy on the validation data as the architecture
has been trained partially on a certain number of training batches while inheriting its weights
from the previous generation.

\subsection{Evolutionary Algorithm}
The evolutionary algorithm (EA) starts with a population of architectures, which are sampled
from a uniform distribution on the interval $\left[0,1\right)$, and it runs for \textit{G}
generations. In each generation, the one shot model is trained on the training data by using the
decoded architecture parameter $\bar{\alpha}$ of each individual architecture in the population in
a round-robin fashion. Then, the fitness of each individual architecture is estimated using the
decoded architecture parameter $\bar{\alpha}$. The population is then evolved using crossover and
mutation operations to create the next generation population replacing the previous generation
population. The best architecture in each generation does not undergo any modification and is
automatically copied to the next generation. This ensures that the algorithm does not forget the
best architecture learned thus far and gives an opportunity to old generation architecture to
compete against the new generation architecture; this is known as \textit{elitism}. The best
architecture is returned after \textit{G} generations. The entire process is summarized as
Algorithm 1 in the supplementary.

\begin{figure}[h]
  \begin{center}
       \includegraphics[width=0.9\linewidth]{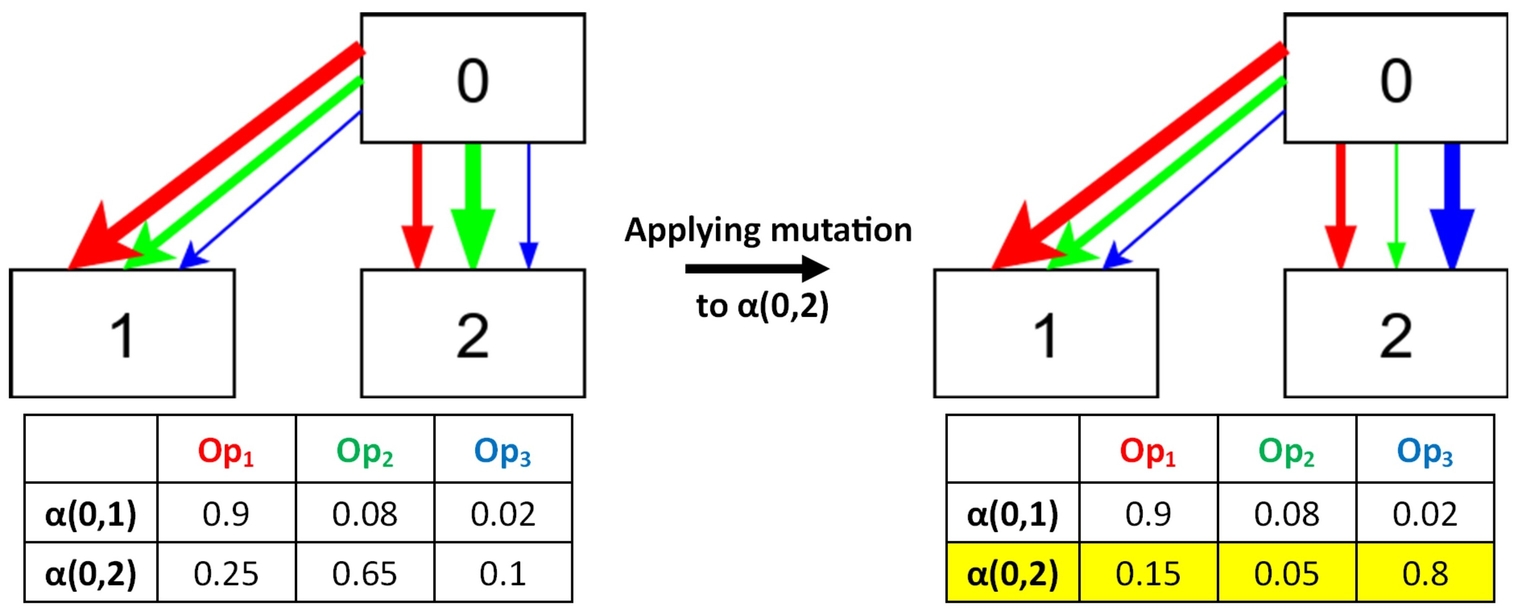}
  \end{center}
  \caption{Illustration of mutation operation}
  \label{fig:mutation}
  \end{figure}

\textbf{Mutation Operation:} It refers to a random change to an individual
architecture in the population. The algorithm uses the \textit{mutation rate}
\cite{eiben2003introduction}, which decides the probability of changing the
architecture parameter, $\alpha^{i,j}$, between
node $i$ and node $j$. This is done by re-sampling $\alpha^{i,j}$ from a 
uniform distribution on the interval $\left[0,1\right)$ as illustrated in 
Figure ~\ref{fig:mutation}. 

\begin{figure*}[t]
  \centering
  \begin{subfigure}{0.45\linewidth}
    \includegraphics[width=\linewidth]{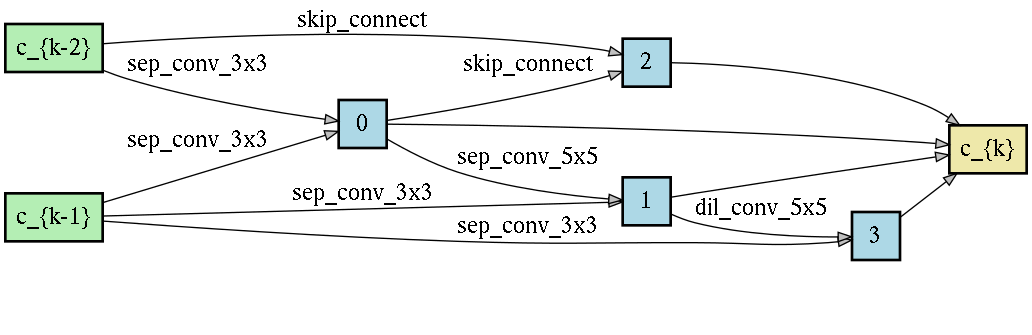}
    \caption{}
    \label{fig:normal}
  \end{subfigure}
  \quad \quad
  \begin{subfigure}{0.45\linewidth}
    \includegraphics[width=\linewidth]{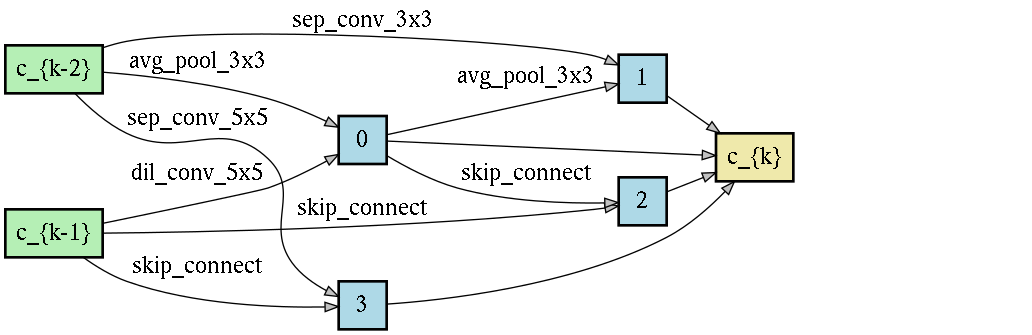}
    \caption{}
    \label{fig:reduce}
  \end{subfigure}
  \caption{Discovered cell using EvNAS-A (a) Normal Cell (b) Reduction Cell}
  \label{discovered_cells}
\end{figure*}

\begin{figure}[h]
  \begin{center}
       \includegraphics[width=\linewidth]{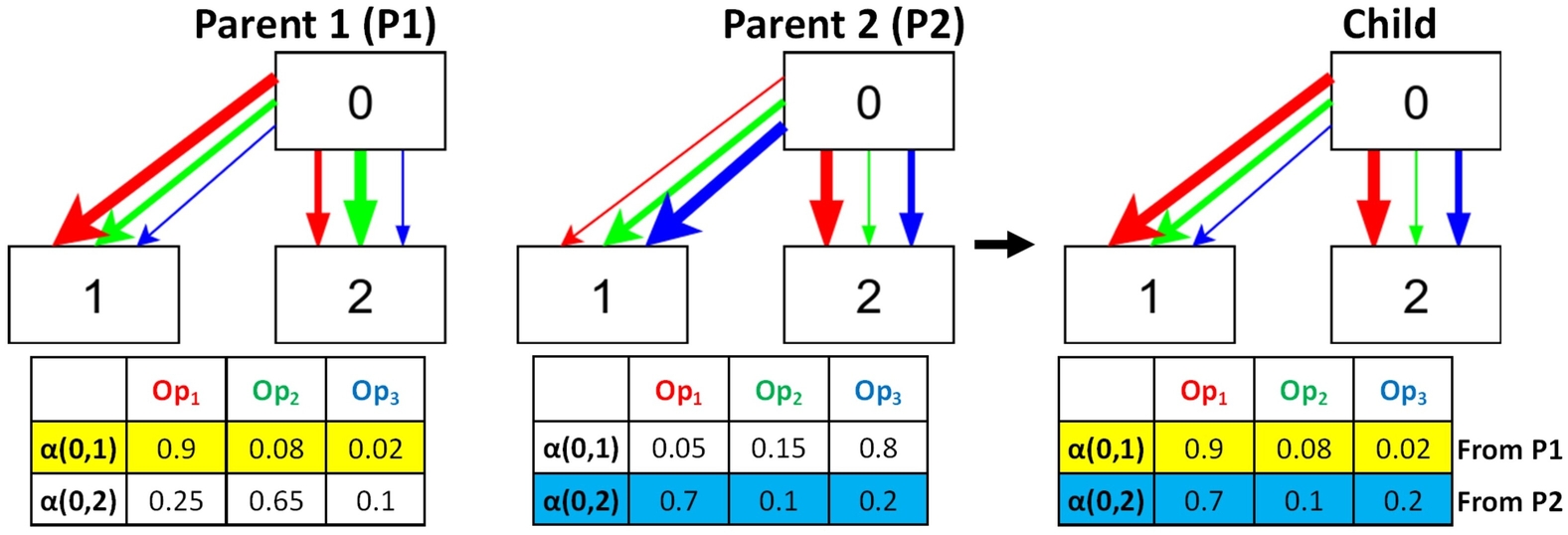}
  \end{center}
  \caption{Illustration of crossover operation}
  \label{fig:crossover}
  \end{figure}

\textbf{Crossover Operation:} It is a process of combining parent architectures
to create a new child architecture, which may perform better than the parents.
EvNAS uses \textit{tournament selection} \cite{eiben2003introduction} for the parent
selection process to generate the next generation architecture population. In
\textit{tournament selection}, a certain number of architectures are randomly selected
from the current population. The top-2 most fit architectures from the 
selected group become \textit{parent1} and \textit{parent2} and are used to create 
a single child architecture. This is done by copying the architecture 
parameters, $[{\alpha}^{i,j}]_{parent1}$ and $[{\alpha}^{i,j}]_{parent2}$ between node $i$ and
node $j$, from \textit{parent1} and \textit{parent2}, respectively, with a certain probability
to the child architecture parameter, $[{\alpha}^{i,j}]_{child}$ between node $i$ and node $j$
as illustrated in Figure ~\ref{fig:crossover}. This can be formulated as follows:
    \begin{equation}
        [{\alpha}^{i,j}]_{child} = 
            \begin{cases}
            [{\alpha}^{i,j}]_{parent1}, \text{with probability 0.5}
            \\
            [{\alpha}^{i,j}]_{parent2},  \text{otherwise}
            \\
            \end{cases}    
    \end{equation}
Note that as all the architectures are sub-graph of the super-graph, i.e. the one shot
model, so, we do not have to keep track of the ancestors in order to apply the crossover
operation, as was done in \cite{zhu2019eena}\cite{stanley2002evolving}.

\subsection{Relationship to Prior Works}
Weight inheritance was used in \cite{real2017large} during architecture search
using evolution but in the proposed method, the architectures in a given generation
share weights and inherit the weights from the previous generation (i.e weight
inheritance) because of the use of the one shot model.

FairNAS \cite{chu2019fairnas}, NSGANetV2 \cite{lu2020nsganetv2} has also proposed evolutionary search
with weight sharing which has two steps for searching neural architecture where they optimizes the
supernet in the first step and then FairNAS performs architecture search using evolutionary method with
the trained supernet as the evaluator in the second step while NSGANetV2 uses the weights from trained
supernet to warm start gradient descent for an architecture during the architecture search. In
contrast, our method combines both the training and search process in one single stage.
FairNAS and NSGANetV2 solves the search problem as a multi-objective problem whereas our method solves
it as a single objective problem.

\section{Experiments and Results}
\label{experiments}
In this section, we report the performance of the proposed algorithm EvNAS in terms of a neural 
architecture search on the CIFAR-10 dataset \cite{krizhevsky2009learning} and the performance of
the found architectures on the CIFAR-100 dataset \cite{krizhevsky2009learning} and the
ImageNet dataset \cite{imagenet_cvpr09}. We then present an ablation study showing the importance
of the proposed \textit{decoded architecture} parameter, $\bar{\alpha}$, crossover and
mutation operations during the search process.

\textbf{Initialization:} Each architecture in a population is represented by the architecture
parameter, $\alpha$, which is sampled from a uniform distribution on the interval
$\left[0,1\right)$.

\textbf{Search Process:} The search process on the CIFAR-10 is divided into three stages as was done in
\cite{liu2018darts2}\cite{li2019random}. In \textit{stage 1}, we perform the search process for a
cell block on CIFAR-10 by using four different seeds; this can be thought of as the search stage
of the algorithm. In \textit{stage 2}, the best architecture found in each trial of stage 1 is
evaluated by retraining a larger network created using the same cell blocks discovered in stage 1
for 600 epochs from scratch on CIFAR-10. Next, we choose the best performing architecture among
the four trials, making it the selection stage of the algorithm. In \textit{stage 3}, we evaluate
the best architecture found from stage 2 by training the network from scratch with ten different
seeds for 600 epochs. This stage can be considered as the evaluation stage of the algorithm.

\begin{table*}[t]
    \caption{Comparison of EvNAS with other NAS methods on CIFAR-10 and CIFAR-100 datasets. The
    first block presents the performance of the hand-crafted architecture. The second block
    presents the performance of other NAS methods, the third block presents the performance of our
    method and the last block presents the performance of our ablation study. All the architecture
    search were performed using cutout. $\dagger$ indicates that the result was reported in
    \cite{chen2019progressive}}
    \label{table:CIFAR10}
    \centering
    \begin{tabular}{lcccccc}
    \hline
     & \multicolumn{2}{c}{\bf{Test Error (\%)}} & \bf{Params} & \bf{Search Time} &\bf{Search} \\
    \bf{Architecture} & \bf{C10} & \bf{C100} & (M) & (GPU Days) & \bf{Method} \\
    \hline
    DenseNet-BC         \cite{huang2017densely}  & 3.46 & 17.18   & 25.6 & - & manual\\
    \hline
    PNAS            \cite{liu2018progressive}  &3.41 & - & 3.2 & 225 & SMBO\\
    NASNet-A   \cite{zoph2018learning}     & 2.65 & -     & 3.3  &1800& RL\\
    ENAS       \cite{pmlr-v80-pham18a}    & 2.86 & -       & 4.6  &0.45& RL\\
    DARTS      \cite{liu2018darts2} & $2.76\pm0.09$ & $17.54^{\dagger}$ & 3.3 &4& gradient-based\\
    SNAS       \cite{xie2018snas}          & $2.85\pm0.02$ & - & 2.8&1.5& gradient-based\\
    PDARTS     \cite{chen2019progressive}  & $2.50$ & 16.55        & 3.4 & 0.3 & gradient-based\\
    AmoebaNet-A\cite{real2019regularized}  & $3.34\pm0.06$ & - & 3.2 &3150& evolution\\
    AmoebaNet-B\cite{real2019regularized}  & $2.55\pm0.05$ & - & 2.8 &3150& evolution\\
    EENA       \cite{zhu2019eena}          & $2.56$ & 17.71 & 8.47 & 0.65 & evolution\\
    Random Search WS\cite{li2019random}         & $2.86\pm0.08$ & - & 4.3 &2.7& random\\
    \hline
    EvNAS-A (Ours)                       & $2.47\pm0.06$ & 16.37 & 3.6 & 4.4 & evolution\\
    EvNAS-B (Ours)                       & $2.62\pm0.06$ & 16.51 & 3.8 & 4.4 & evolution\\
    EvNAS-C (Ours)                       & $2.63\pm0.05$ & 16.86 & 3.4 & 4.4 & evolution\\
    \hline
    EvNAS-Rand (Ours)                       & $2.84\pm0.08$ & - & 2.7 &0.62& random\\
    
    EvNAS-ND (Ours)                   & $2.78\pm0.1$ & - & 3.8 & 4.4 & evolution\\
    EvNAS-NDF (Ours)                   & $2.75\pm0.09$ & - & 3.1 & 4.4 & evolution\\
    EvNAS-NDT (Ours)                   & $2.67\pm0.06$ & - & 3.5 & 4.4 & evolution\\
    EvNAS-Mut (Ours)                   & $2.79\pm0.06$ & - & 3.4 & 4.4 & evolution\\
    EvNAS-Cross (Ours)                   & $2.81\pm0.08$ & - & 3.2 & 4.4 & evolution\\
    
    \end{tabular}
\end{table*}

\subsection{Search on CIFAR-10:} \label{res:cifar10}
\textbf{Dataset:} CIFAR-10 \cite{krizhevsky2009learning} has 50,000 training
images and 10,000 testing images with a fixed resolution of 32x32. During
the architecture search, the training images are divided into two subsets of size
25,000 each, out of which the first subset is used for training the one shot model
and the other subset is the validation data, which is used for calculating the \textit{fitness}
of each architecture in the population. In the selection and evaluation stage, the normal
training/testing split is used.

\textbf{Search Space:} We follow the setup given in DARTS\cite{liu2018darts2}. The one shot model is
created by stacking \textit{normal} cell with \textit{reduction} cell inserted at 1/3 and 2/3 of the
total depth of the model. Each cell has two input nodes, four intermediate node and one output
node resulting in 14 edges among them. The operations considered for the cells are as follows: 
3x3 and 5x5 dilated separable convolutions, 3x3 and 5x5 separable convolutions, 3x3 max pooling, 
3x3 average pooling, skip connect and zero. Thus, each architecture is represented by two
14x8 matrices one for normal cell and one for reduction cell.

\begin{table*}[t]
    \caption{Comparison of our method with other image classifiers on ImageNet in mobile setting.
    The first block presents the performance of the hand-crafted architecture. The second block
    presents the performance of other NAS methods and the last block presents the performance of our
    method.}
    \label{table:imagenet}
    \centering
    \begin{tabular}{lcccccc}
    \hline
     &  \multicolumn{2}{c}{\bf{Test Error (\%)}} & \bf{Params} &+$\times$& \bf{Search Time} &\bf{Search} \\
    \bf{Architecture} & \bf{top 1} & \bf{top 5} & (M) & (M) & (GPU Days) & \bf{Method} \\
    \hline
    MobileNet            \cite{howard2017mobilenets}&29.4& 10.5    & 4.2 & 569 & - & manual\\ 
    \hline
    PNAS            \cite{liu2018progressive}  &25.8& 8.1    & 5.1 & 588 & 225 & SMBO\\
    NASNet-A             \cite{zoph2018learning}    & 26.0 & 8.4      & 5.3 & 564 &1800& RL\\
    NASNet-B             \cite{zoph2018learning}    & 27.2  & 8.7     & 5.3 & 488 &1800& RL\\
    NASNet-C             \cite{zoph2018learning}    & 27.5  & 9.0     & 4.9 & 558 &1800& RL\\
    
    DARTS  \cite{liu2018darts2}       & 26.7  & 8.7 & 4.7 & 574 &4& gradient-based\\
    SNAS                 \cite{xie2018snas}         & 27.3  & 9.2 & 4.3 & 522 &1.5& gradient-based\\
    PDARTS               \cite{chen2019progressive} & 24.4  & 7.4 & 4.9 & 557 &0.3& gradient-based\\
    
    AmoebaNet-A          \cite{real2019regularized} & 25.5 & 8.0 & 5.1 & 555 &3150& evolution\\
    AmoebaNet-B          \cite{real2019regularized} & 26.0 & 8.5 & 5.3 & 555 &3150& evolution\\
    AmoebaNet-C          \cite{real2019regularized} & 24.3 & 7.6 & 6.4 & 570 &3150& evolution\\
    FairNAS-A          \cite{chu2019fairnas} & 24.7 & 7.6 & 4.6 & 388 &12& evolution\\
    FairNAS-B          \cite{chu2019fairnas} & 24.9 & 7.7 & 4.5 & 345 &12& evolution\\
    FairNAS-C          \cite{chu2019fairnas} & 25.3 & 7.9 & 4.4 & 321 &12& evolution\\
    \hline
    EvNAS-A (Ours)                                     & 24.4 & 7.4 & 5.1 & 570 & 4.4 & evolution\\
    EvNAS-B (Ours)                                     & 24.4 & 7.4 & 5.3 & 599 & 4.4 & evolution\\
    EvNAS-C (Ours)                                     & 25.1 & 7.8 & 4.9 & 547 & 4.4 & evolution\\
    
    \end{tabular}
\end{table*}

\textbf{Training Settings:} The training setting mainly follows the setup proposed by DARTS
\cite{liu2018darts2}. Because of the high memory requirements of the one shot model, a smaller
network, called \textit{proxy network} \cite{li2019random}, with 8 stacked cells and 
16 initial channels is used during the architecture search process, i.e. \textit{stage 1}.
For deriving the discrete architecture, $arch_{dis}$, each node in the discrete architecture
is connected to two nodes among the previous nodes selected via the top-2 operations
according to the architecture parameter $\alpha$. During the search process, we use SGD for
training the one shot model with a batch size of 64, initial learning rate of 0.025, momentum of
0.9, and weight decay of $3\times10^{-4}$.  The learning rate is annealed down to 0.001 by using
the cosine annealing schedule without any restart during the search process. For our evolutionary
algorithm, we use a population size of 50 in each generation, 0.1 as the \textit{mutation rate}
and 10 architectures are chosen randomly during the tournament selection. The search process runs
for 50 generations on a single GPU, NVIDIA 2080 Ti, and takes 4.4 days to complete stage 1. Number
of generations was chosen to match the number of epochs in DARTS \cite{liu2018darts2}. Population
size was chosen based on the experiments where we ran our method for population size of 20, 30, 50
with tournament size chosen as one-fifth of the population size, as shown in Figure
\ref{fig:exp}(b). We did not go beyond 50 population size as we wanted to have search time
similar to that of DARTS. Mutation rate was chosen based on the experiments where we ran our
method for \textit{mutation rate} of 0.05, 0.1, 0.15, as shown in Figure \ref{fig:exp}(c). All our
architecture search in Table \ref{table:CIFAR10} are done with cutout \cite{devries2017improved}.

\textbf{Architecture Evaluation:} A larger network, called \textit{proxyless network} 
\cite{li2019random}, with 20 stacked cells and 36 initial channels is used during the selection
and evaluation stage. Following DARTS\cite{liu2018darts2}, the proxyless network is trained with a
batch size of 96, weight decay of 0.0003, cutout \cite{devries2017improved}, auxiliary tower with
0.4 as its weights, and path dropout probability of 0.2 for 600 epochs. The same setting is used
to train and evaluate the proxyless network on the CIFAR-100 dataset
\cite{krizhevsky2009learning}.

\textbf{Search Results and Transferability to CIFAR-100:} We perform the architecture search on
CIFAR-10 three times with different random number seeds and their
results are provided in Table~\ref{table:CIFAR10} as EvNAS-A, EvNAS-B and EvNAS-C, which are then
transferred to CIFAR-100. The cells discovered during EvNAS-A are shown in
Figure~\ref{discovered_cells} and those discovered by EvNAS-B and EvNAS-C are given in the
supplementary. EvNAS-A evaluates 10K architectures during the search time and achieves the
average test error of $2.47\pm0.06$ and $16.37$ on CIFAR-10 and CIFAR-100 respectively with
search time significantly less than the previous evolution based methods.
EENA \cite{zhu2019eena} found a competitive architecture in lesser search time than EvNAS using
evolution but EvNAS was able to achieve better result on both CIFAR-10 and CIFAR-100 with fewer
parameters. 

\subsection{Architecture Transferability to ImageNet:}
\textbf{Architecture Evaluation:} The architecture discovered in the search process on CIFAR-10 is
then used to train a network on the ImageNet dataset \cite{imagenet_cvpr09} with 14 cells and 48
initial channels in the mobile setting, where the size of the input images is 224 x 224 and the
number of multiply-add operations in the model is restricted to less than 600M. We follow the
training settings used by PDARTS \cite{chen2019progressive}. The network is trained from scratch with a batch size of 1024 on 8 NVIDIA V100 GPUs.

\textbf{ImageNet Results:}  The results of the evaluation on the ImageNet dataset are provided in
Table ~\ref{table:imagenet}. The result shows that the cell discovered by EvNAS on CIFAR-10 can be
successfully transferred to the ImageNet, achieving a top-5 error of 7.4\%. Notably, EvNAS is able
to achieve better result than previous state-of-the-art evolution based methods AmoebaNet
\cite{real2019regularized}, FairNAS \cite{chu2019fairnas} while using significantly less computational resources.

\subsection{Ablation Studies} \label{ablation}
To discover the effect of the \textit{decoded architecture} parameter, $\bar{\alpha}$, and the
crossover and mutation operations during the search process, we conduct more architecture
searches: without \textit{decoded architecture} parameter, $\bar{\alpha}$, with \textit{crossover 
only}, with \textit{mutation only} and without \textit{crossover and mutation}. The search results
are provided in Table~\ref{table:CIFAR10}.

\textbf{Without \textit{Crossover and Mutation}:} Here, a population of 50 architectures are
randomly changed after every generation and in the last generation, the architecture with the best
performance on the validation set is chosen as the best found architecture. Thus, the search
process only evaluates only 200 architectures to come up with the best architecture. The
architecture found (listed as \textit{EvNAS-Rand} in Table~\ref{table:CIFAR10}) achieves an average
error of $2.84\pm0.08$, as the search behaves as a \textit{random search} and shows similar results
to those reported in \cite{li2019random}.

\textbf{Without \textit{Decoded Architecture} Parameter, $\bar{\alpha}$:} Here, we conduct three
architecture searches where a population of 50 architectures are modified through both crossover
and mutation operations without using the decoded architecture parameter, $\bar{\alpha}$, (i)
during the training (\textit{EvNAS-NDT}), (ii) during the fitness evaluation of each individual
architecture in the population (\textit{EvNAS-NDF}) and (iii) during both training and fitness
evaluation (\textit{EvNAS-ND}). The architecture found (listed in Table~\ref{table:CIFAR10})
in \textit{EvNAS-NDT} performs better than that of \textit{EvNAS-NDF} which shows that the decoded
architecture parameter, $\bar{\alpha}$, is more important during the fitness estimation step
than during the training step. Also, the architecture found in \textit{EvNAS-ND} performs slightly
better than that of the \textit{random search} because of the direction component introduced by the
crossover operation. The improvement is due to the fact that when using architecture
parameter, $\alpha$, it allows a varying amount of contribution from other architectures during the
fitness estimation of a particular architecture from the one shot model, resulting in very noisy
fitness estimate. But when using the decoded architecture parameter, $\bar{\alpha}$, it assigns
higher weight to the current architecture while giving equally small weights to other architectures.

\textbf{With \textit{Mutation Only}:} Here, a population of 50 architectures are modified only
through a mutation operation with 0.1 as the \textit{mutation rate} while using the decoded
architecture parameter. The architecture found (listed as
\textit{EvNAS-Mut} in Table~\ref{table:CIFAR10}) performs slightly better than that of the
\textit{random search} even though mutation is a random process. This improvement can be attributed
to \textit{elitism}, which does not let the algorithm forget the best architecture learned thus far. 

\textbf{With \textit{Crossover Only}:} Here, a population of 50 architectures are
modified only through a crossover operation only while using the decoded architecture parameter.
The architecture found (listed as \textit{EvNAS-Cross} in Table~\ref{table:CIFAR10})
performs slightly better than that of the \textit{random search}. This improvement can be attributed
to the selection pressure \cite{eiben2003introduction} introduced because of the tournament
selection, which guides the search towards the better architecture solution. 

The improvements in both \textit{EvNAS-Mut} and \textit{EvNAS-Cross} are not much as compared to
the \textit{EvNAS-Rand} because of the fact that we are using a partially trained network for
evaluating the architectures on the validation set which provides a noisy estimate of their
fitness/performance. The ablation study shows that the decoded architecture parameter
$\bar{\alpha}$, mutation and crossover operations play an equally important role in the search process while the decoded
architecture parameter, $\bar{\alpha}$, plays more important role during the fitness estimation.
All the cells discovered in the ablation study are provided in the supplementary.

\subsection{Discussion on Evolutionary Search vs Gradient Based Search}
The gradient based methods are highly dependent on the search space and they tend to overfit to
operations that lead to faster gradient descent which is the \textit{skip-connect} operation due
to its parameter-less nature leading to higher number of \textit{skip-connect} in the final
discovered cell \cite{Zela2020Understanding}. PDARTS \cite{chen2019progressive} uses a
regularization method to restrict the number of \textit{skip-connect} to a specific number in the
final normal cell for the search space used in the original DARTS paper. PDARTS emperically found
that the optimal number of \textit{skip-connect} in the normal cell is 2, which is a search space
dependent value and the optimal number may not be 2 if the search space is changed. This reduces the
search space resulting in faster search time as compared to the original DARTS which is a gradient
based method without any regularization applied to the search space. Notice that without  such
regularization to restrict the number of skip-connect to 2, the gradient based methods, e.g. DARTS,
only provides similar search time but much worse performance than ours due to the overfitting problem.
By contrast, EvNAS does not have to worry about the overfitting problem due to its stochastic nature
and so it is not dependent on the search space. EvNAS arrives at this optimal solution without any
regularization being applied to the search space as can be seen in the discovered normal cells in
Figure \ref{discovered_cells}(a) and all the figures in the supplementary.

\section{Conclusions and Future Directions}
We propose an efficient method of applying a simple genetic algorithm to the
neural architecture search problem with both parameter sharing among the individual
architectures and weight inheritance using a one shot model, resulting in decreased
computational requirements as compared to other evolutionary search methods.
A decoding method for the architecture parameter was used to improve the fitness
estimation of a partially trained individual architecture from the one shot model.
The proposed crossover along with the tournament selection provides
a direction to an otherwise directionless random search. The proposed algorithm
was able to significantly reduce the search time of evolution based architecture search
while achieving better results on CIFAR-10, CIFAR-100 and ImageNet dataset than previous
evolutionary algorithms. A possible future direction to improve the
performance of the algorithm is by making an age factor to be a part of the architecture,
which makes sure that the old generation architectures do not die after one generation
and can compete against the newer generation architectures.

{\small
\bibliographystyle{ieee_fullname}
\bibliography{egbib}
}

\end{document}


\title{Supplementary}

\maketitle
\begin{abstract}
Here, we present the summary of the algorithm and all the discovered cells from EvNAS-B, EvNAS-C and
the ablation study which are shown in Figure~\ref{fig:EvNASB} to Figure~\ref{fig:cross_cells}.
\end{abstract}

\section{Algorithm}
\begin{algorithm}[h]
    \begin{algorithmic}
        \STATE \textbf{Input:} population \textit{P}, population size \textit{N}, 
        number of selected individuals in tournament selection \textit{T}, 
        mutation rate \textit{r}, total number of training batches \textit{B}, 
        one shot model \textit{M}
        \FOR{\textit{g} = 1, 2, ..., \textit{G} generations}
            \FOR{\textit{i} = 1,..., \textit{B}} 
                \STATE Copy $\bar{\alpha}[i \mod N]$ generated from 
                $\alpha[i\mod N]$ to \textit{M} and train \textit{M} on training batch[\textit{i}];
            \ENDFOR
            \STATE Calculate the fitness of each individual architecture, $\alpha$ in the population
            by copying the respective $\bar{\alpha}$ to \textit{M};
            \STATE Set Elite, \textit{E} $\gets$ best architecture in \textit{P};
            \STATE Copy \textit{E} to the next generation population, \textit{$P_{next}$};
            \FOR{\textit{i} = 2,.., \textit{N}}
                \STATE Select \textit{T} individuals randomly from \textit{P} \{Tournament
                Selection\} and use top-2 individuals to create new architecture using
                \textit{crossover} operation for \textit{$P_{next}$};
                \FOR{\textit{j} = 1,.., length($\alpha[i]$)}
                    \IF{$uniformRandom(0,1) \leq r$} 
                        \STATE Apply \textit{mutation} operation to the $j^{th}$ component of
                        $\alpha[i]$ in \textit{$P_{next}$};
                     \ENDIF
                 \ENDFOR
            \ENDFOR
            
            \STATE \textit{$P$} $\gets$ \textit{$P_{next}$};
        \ENDFOR
        \RETURN Elite, \textit{E}
    \end{algorithmic}
    \caption{EvNAS}
    \label{algo:ENAS_WS}
\end{algorithm}

Where $\alpha[i]$ represents the $i^{th}$ architecture in the population.

\section{Discovered Cells from EvNAS-B, EvNAS-C and Ablation Study}
\begin{figure}[H]
  \centering
  \begin{subfigure}{1.0\linewidth}
    \includegraphics[width=\linewidth]{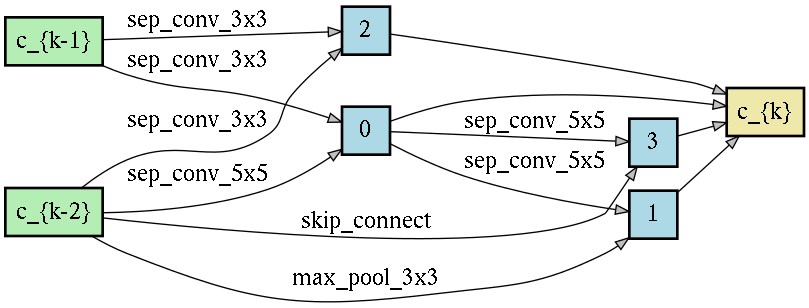}
    \caption{}
  \end{subfigure}
  \quad \quad
  \begin{subfigure}{1.0\linewidth}
    \includegraphics[width=\linewidth]{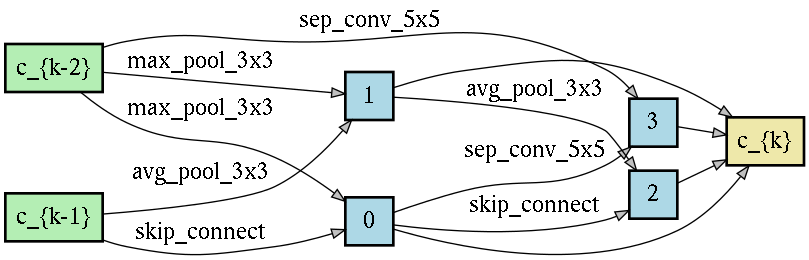}
    \caption{}
  \end{subfigure}
  \caption{Discovered cell in EvNAS-B (a) Normal Cell (b) Reduction Cell.}
  \label{fig:EvNASB}
\end{figure}

\begin{figure}[t]
  \centering
  \begin{subfigure}{1.0\linewidth}
    \includegraphics[width=\linewidth]{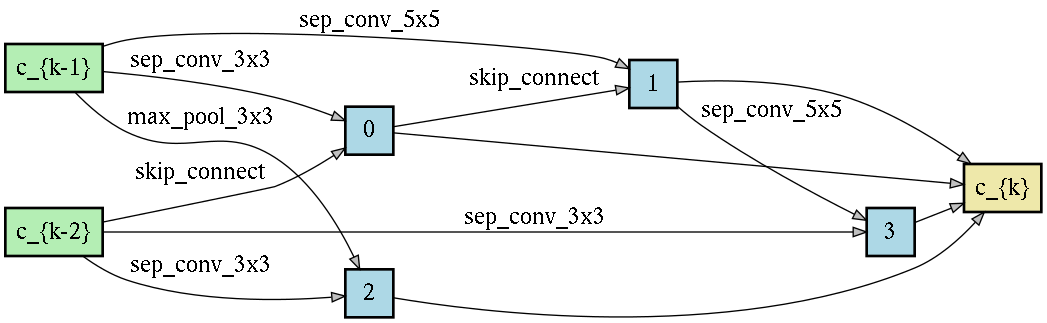}
    \caption{}
  \end{subfigure}
  \quad \quad
  \begin{subfigure}{1.0\linewidth}
    \includegraphics[width=\linewidth]{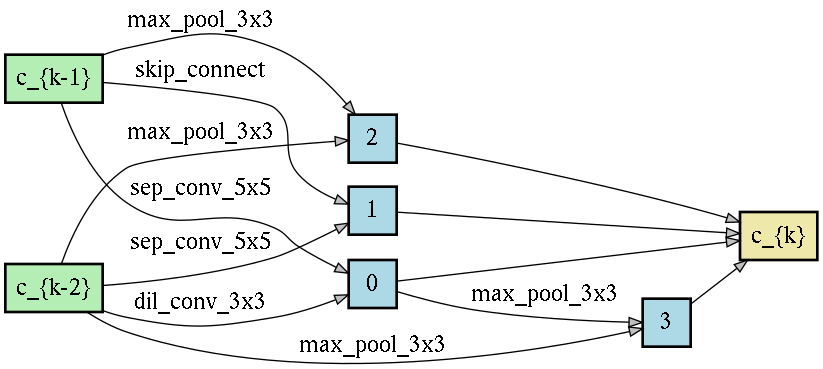}
    \caption{}
  \end{subfigure}
  \caption{Discovered cell in EvNAS-C (a) Normal Cell (b) Reduction Cell.}
  \label{fig:EvNASC}
\end{figure}

\begin{figure}[t]
  \centering
  \begin{subfigure}{1.0\linewidth}
    \includegraphics[width=\linewidth]{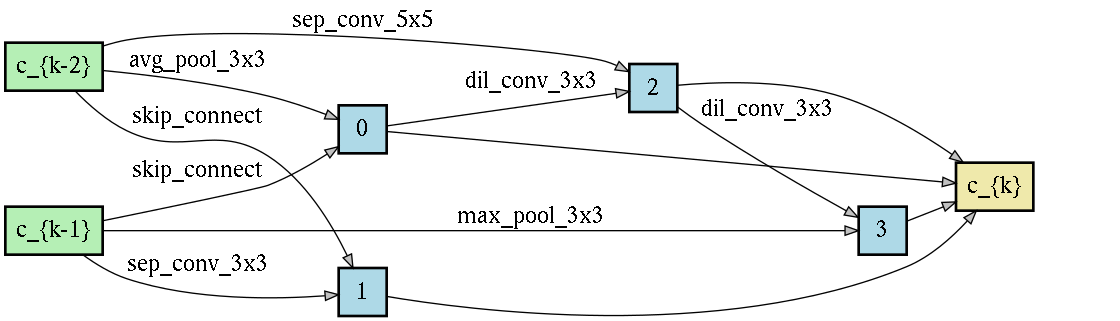}
    \caption{}
  \end{subfigure}
  \quad \quad
  \begin{subfigure}{1.0\linewidth}
    \includegraphics[width=\linewidth]{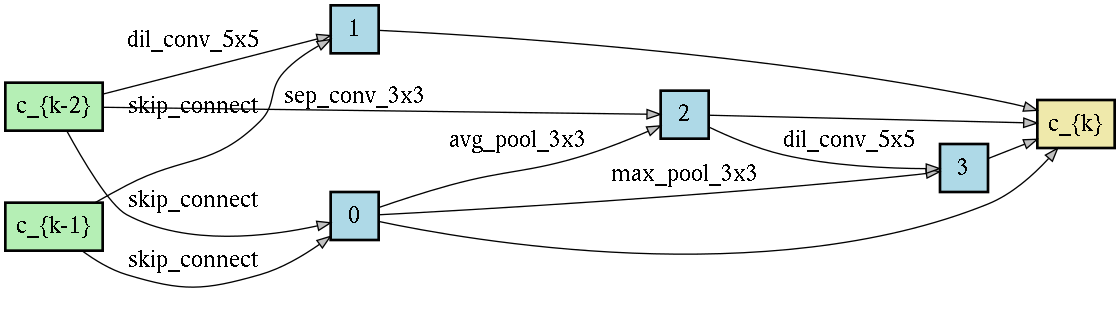}
    \caption{}
  \end{subfigure}
  \caption{Discovered cell using random search (EvNAS-Rand) (a) Normal Cell (b) Reduction Cell.}
  \label{fig:random_cells}
\end{figure}

\begin{figure}[h]
  \centering
  \begin{subfigure}{1.0\linewidth}
    \includegraphics[width=\linewidth]{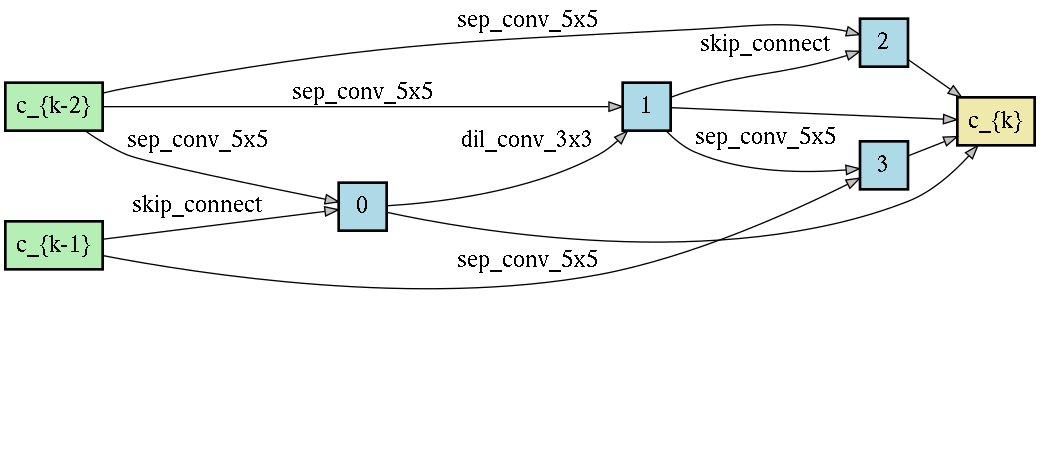}
    \caption{}
  \end{subfigure}
  \quad \quad
  \begin{subfigure}{1.0\linewidth}
    \includegraphics[width=\linewidth]{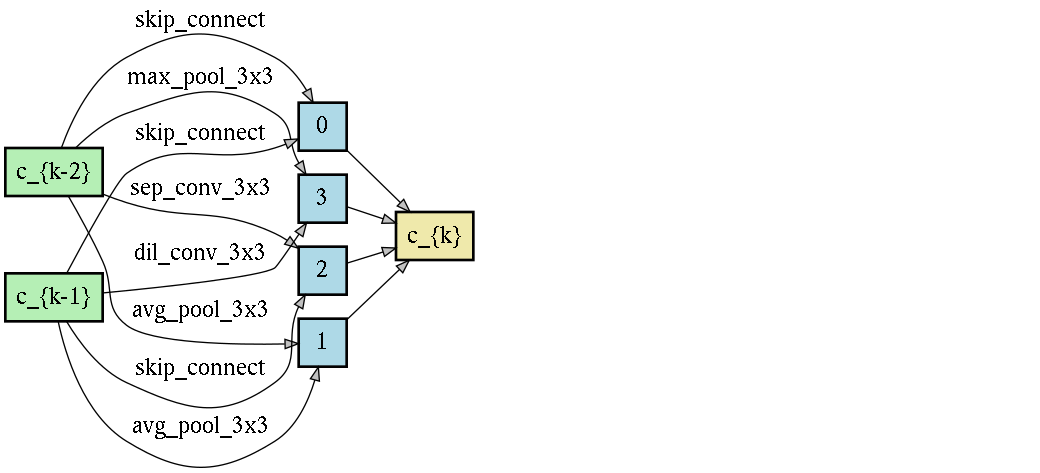}
    \caption{}
  \end{subfigure}
  \caption{Discovered cell using EvNAS without decoding architecture $\bar{\alpha}$ during both
  training and fitness evaluation (EvNAS-ND) (a) Normal Cell (b) Reduction Cell.}
  \label{fig:no_dis_cells}
\end{figure}

\begin{figure}[t]
  \centering
  \begin{subfigure}{1.0\linewidth}
    \includegraphics[width=\linewidth]{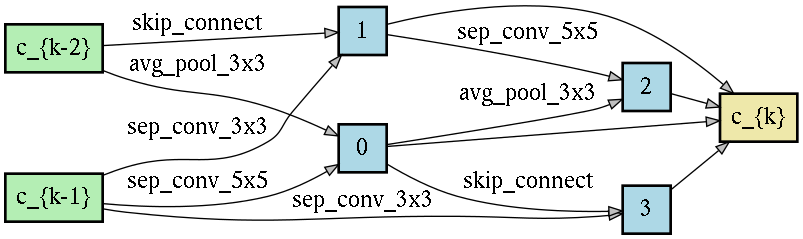}
    \caption{}
  \end{subfigure}
  \quad \quad
  \begin{subfigure}{1.0\linewidth}
    \includegraphics[width=\linewidth]{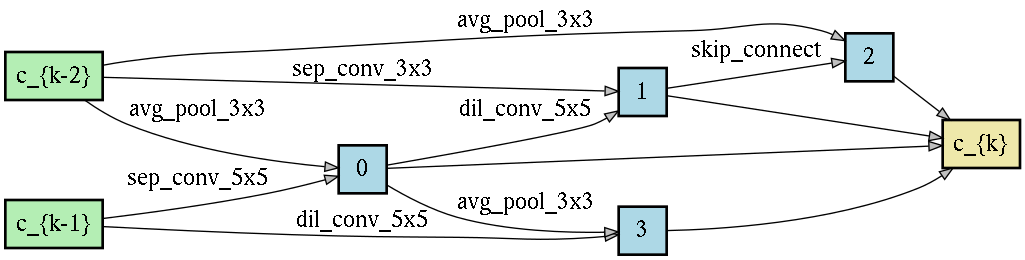}
    \caption{}
  \end{subfigure}
  \caption{Discovered cell using EvNAS without decoding architecture $\bar{\alpha}$ during fitness evaluation (EvNAS-NDF) (a) Normal Cell (b) Reduction Cell.}
  \label{fig:EvNAS-NDF}
\end{figure}

\begin{figure}[h]
  \centering
  \begin{subfigure}{1.0\linewidth}
    \includegraphics[width=\linewidth]{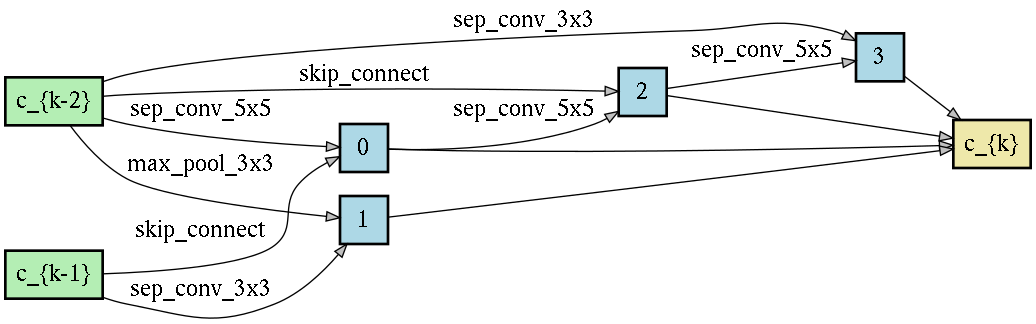}
    \caption{}
  \end{subfigure}
  \quad \quad
  \begin{subfigure}{1.0\linewidth}
    \includegraphics[width=\linewidth]{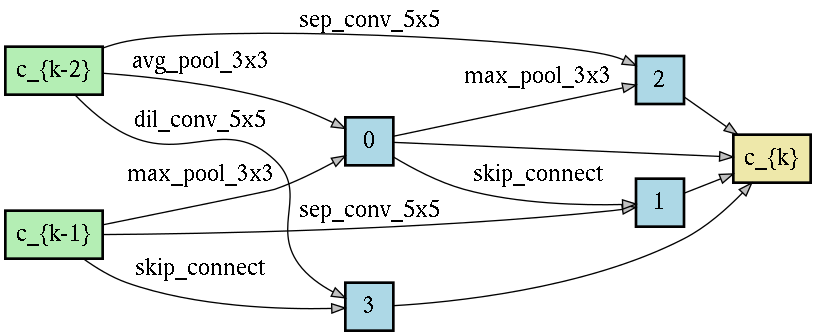}
    \caption{}
  \end{subfigure}
  \caption{Discovered cell using EvNAS without decoding architecture $\bar{\alpha}$ during training (EvNAS-NDT) (a) Normal Cell (b) Reduction Cell.}
  \label{fig:EvNAS-NDT}
\end{figure}

\begin{figure}[h]
  \centering
  \begin{subfigure}{1.0\linewidth}
    \includegraphics[width=\textwidth]{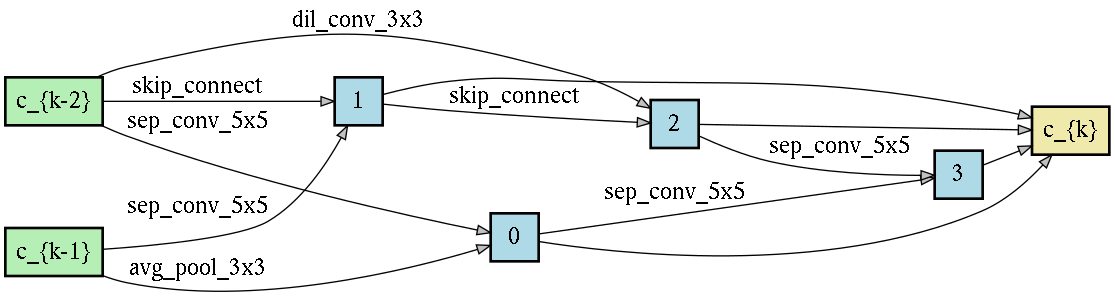}
    \caption{}
  \end{subfigure}
  \quad \quad
  \begin{subfigure}{1.0\linewidth}
    \includegraphics[width=\textwidth]{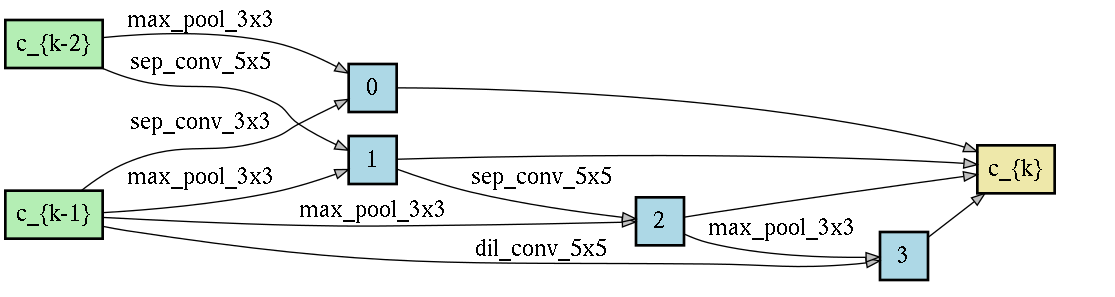}
    \caption{}
  \end{subfigure}
  \caption{Discovered cell using EvNAS with mutation only (EvNAS-Mut) (a) Normal Cell (b) Reduction Cell.}
  \label{fig:mut_cells}
\end{figure}

\begin{figure}[t]
  \centering
  \begin{subfigure}{1.0\linewidth}
    \includegraphics[width=\linewidth]{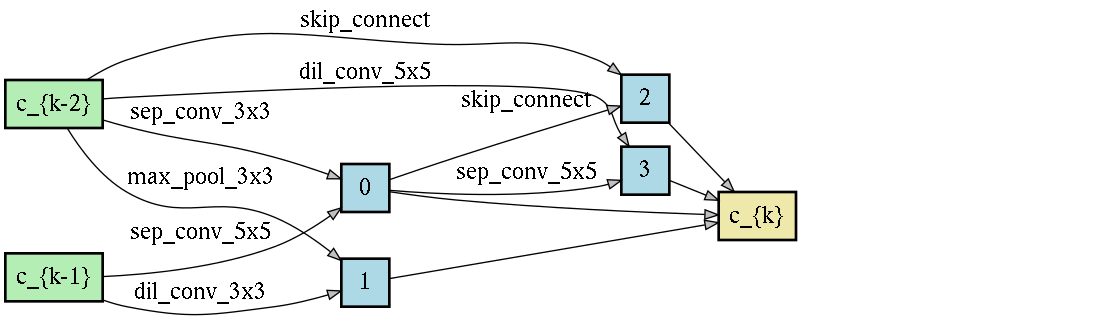}
    \caption{}
  \end{subfigure}
  \quad \quad
  \begin{subfigure}{1.0\linewidth}
    \includegraphics[width=\linewidth]{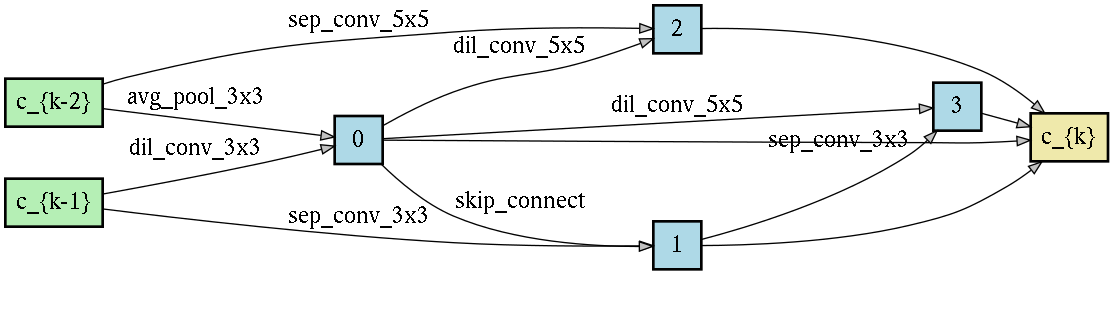}
    \caption{}
  \end{subfigure}
  \caption{Discovered cell using EvNAS with crossover only (EvNAS-Cross) (a) Normal Cell (b) Reduction Cell.}
  \label{fig:cross_cells}
\end{figure}

\bigskip